\documentclass[sigconf]{acmart}
\usepackage{tabularx}
\usepackage{adjustbox}
\usepackage[normalem]{ulem}
\usepackage{multirow}
\usepackage{booktabs}
\usepackage[framemethod=TikZ]{mdframed}
\usepackage{amsmath,amsfonts}
\usepackage{algorithm}
\usepackage{algpseudocode}
\usepackage{booktabs} 
\usepackage{soul}
\usepackage{xcolor}

\AtBeginDocument{%
  \providecommand\BibTeX{{%
    \normalfont B\kern-0.5em{\scshape i\kern-0.25em b}\kern-0.8em\TeX}}}

\copyrightyear{2024}
\acmYear{2024}
\setcopyright{acmlicensed}\acmConference[ICSE-NIER'24]{New Ideas and
Emerging Results }{April 14--20, 2024}{Lisbon, Portugal}
\acmBooktitle{New Ideas and Emerging Results (ICSE-NIER'24), April 14--20,
2024, Lisbon, Portugal}
\acmDOI{10.1145/3639476.3639759}
\acmISBN{979-8-4007-0500-7/24/04}

\acmConference[ICSE 2024]{46th International Conference on Software Engineering}{April 14-20,
  2024}{Lisbon}
\acmSubmissionID{icse2024-nier-p14}

\begin{document}

\title{XAIport: A Service Framework for the Early Adoption of XAI in AI Model Development}


\author{Zerui Wang, Yan Liu, Abishek Arumugam Thiruselvi, Abdelwahab Hamou-Lhadj}
\email{{zerui.wang, abishek.arumugamthiruselvi}@mail.concordia.ca; {yan.liu, wahab.hamou-lhadj}@concordia.ca}
\orcid{0000-0002-0403-3112}
\affiliation{%
  \institution{Concordia University}
  \city{Montreal}
  \country{Canada}
}








\begin{abstract}
In this study, we propose the early adoption of Explainable AI (XAI) with a focus on three properties: 
Quality of explanation, the explanation summaries should be consistent across multiple XAI methods; 
Architectural Compatibility, for effective integration in XAI, the architecture styles of both the XAI methods and the models to be explained must be compatible with the framework;
Configurable operations, XAI explanations are operable, akin to machine learning operations. 
Thus, an explanation for AI models should be reproducible and tractable to be trustworthy. 
We present XAIport, a framework of XAI microservices encapsulated into Open APIs to deliver early explanations as observation for learning model quality assurance. 
XAIport enables configurable XAI operations along with machine learning development. 
We quantify the operational costs of incorporating XAI with three cloud computer vision services on Microsoft Azure Cognitive Services, Google Cloud Vertex AI, and Amazon Rekognition. 
Our findings show comparable operational costs between XAI and traditional machine learning, with XAIport significantly improving both cloud AI model performance and explanation stability.

\end{abstract}





\maketitle

\section{Introduction}

Machine Learning Operations (MLOps) is a multidisciplinary approach that includes a set of best practices, concepts, and developments \cite{kreuzberger2023machine}. 
The major tasks of MLOps include automating the ML lifecycle, such as model development, validation, quality assurance, deployment, monitoring, and governance \cite{testi2022mlops}.

The quality assurance of MLOps involves several key components, such as data validation, feature engineering assessment, model training evaluation, cross-validation, performance metrics analysis, fairness, bias evaluation, and model explainability \cite{studer2021towards}. 
The explainability of models is necessary in sensitive domains. The lack of model explanation leads to distrust in the AI models \cite{sheu2022survey,hopkins2021machine}. 


Post-hoc XAI methods provide explanations for complex, already-trained models. The Post-hoc XAI techniques, such as SHAP \cite{lundberg2017unified}, provide feature attribution as explanations. 
XAI operations are often considered a post-hoc activity \cite{rudin2019stop}, implemented after the model has been trained and verified. The quality assurance \cite{studer2021towards} of complex software development has shown that incremental development and iterative quality control are efficient and cost-effective. Inspired by this principle,  we argue that early adoption of XAI operations enhances the quality assurance of AI models with probing observations at the feature representation level and summarized explanations across datasets and AI models. 

We present \texttt{XAIport}, an XAI service architecture that allows  XAI early adoption across cloud platforms and offers unified open API access.
Inconsistent explanations can be misleading to evaluate the AI models.
The probing results and derived explanations should be quantified for their stability and consistency across datasets and AI models.  
The efficiency of applying XAI operations should be measured quantitatively so that the runtime overhead and cost are well balanced in evaluating the benefits of adopting XAIs. 
We summarize the key considerations for the early adoption of XAI operations as follows:

\begin{itemize}
    \item \textbf{Quality of Explanation.} Explanations generated by XAI methods should adhere to the evaluation metrics, specifically the explanation consistency metrics, defined in an established XAI process \cite{xaiprocess}.
    \item \textbf{Architecture Compatibility.} The XAI service flexibly integrates the AI models and XAI methods within the microservice
    architecture. This ensures incorporation into existing cloud services via open APIs.
    \item \textbf{Cost-Efficiency in CI/CD.} The adoption of XAI into MLOps should result in proportional cost-efficient operational overhead during the Continuous Integration and Continuous Deployment (CI/CD) phases. In ideal scenarios, the additional complexity XAI introduces is approximately proportional to existing MLOps.
\end{itemize}
This approach mirrors best practices in software engineering, where early integration of unit tests and quality assurance solidifies the software.
Additionally, \texttt{XAIport} provides a unified measurement of resource consumption and XAI operation overhead. 

\section{Related Works}
\label{sec:relatedworks}
We explore the increasing importance of XAI in AI domains, such as healthcare and finance. Subsequently, we review the MLOps workflow. Then, we review XAI methods in the field of computer vision and metrics for the assessment. 

Alongside the complexity of AI models, the need for explainability has concurrently risen ~\cite{adadi2018peeking}. 
Explainability fosters a better understanding of model behavior, facilitates trust and encourages responsible AI usage ~\cite{arrieta2020explainable}.
In the critical sectors of
healthcare, finance, and legal systems, XAI is essential
in comprehending the model's decisions and implications and ensuring compliance with legal and ethical protocols~\cite{arrieta2020explainable, pawar2020incorporating, ohana2021explainable}. The healthcare domain witnesses an especially pronounced need for XAI due to the growing reliance on AI technologies~\cite{Lambin2017RadiomicsTB}.

MLOps integrate machine learning and operations, emphasizing the importance of explainability or XAI~\cite{kreuzberger2023machine}. The process begins with defining system requirements~\cite{Habibullah2022NonFunctionalRF}. During data collection, potential biases in the data are often overlooked~\cite{whang2020data}. Data preprocessing techniques, such as cutmix~\cite{cutmix} and puzzlemix~\cite{puzzlemix}, aim to improve dataset quality. Feature engineering is central to ML models, and incorporating explainability during feature selection simplifies the model~\cite{Bengio,zacharias2022designing}. Traditional model quality assurance metrics are expanded to include explainability, especially in cloud AI services~\cite{Zuhair2022AnalysisOB, azure-cognitive-services}. Consistency evaluations in XAI ensure trustworthy explanations~\cite{xaiprocess}. Deployment in MLOps emphasizes the use of visualization tools for better model understanding~\cite{xaiprocess}.

Class Activation Mapping (CAM) \cite{zhou2016learning} emerged as a pioneering approach leveraging the global average pooling layer to localize features within Convolutional Neural Network models. A limitation, however, was its need to adjust the model's fully connected layer. In contrast, Grad-CAM \cite{selvaraju2017grad} refined this by determining the localization weight through the layer's average gradient, eliminating the need for replacements. Advancing this further, Grad-CAM++ \cite{Grad-cam++} incorporated second-order gradients for enhanced precision. EigenCAM \cite{EigenCAM} uniquely uses the primary component of activations without class-specific considerations. LayerCAM~\cite{layercam} assigns spatial weights to activations considering only positive gradients, while XGrad-CAM~\cite{fu2020axiom} adjusts gradients based on normalized activations. Representing the forefront of XAI methodologies, these techniques have proven their prowess in generating saliency maps for visual-based XAI tasks.

The metrics for evaluating explainability in XAI are essential and gain considerable attention \cite{chinu2023explainable}. 
However, the field is still grappling with several challenges. Vilone et al. \cite{vilone2021notions} list scientific papers for approaches to evaluate the XAI method and point out the lack of consensus in defining unified evaluation metrics. 
Instead of qualitative metrics, we prefer quantitative metrics to assess XAI methods concretely. A systematic assessment \cite{xaiprocess} provides clear consistency metrics to XAI feature contributions. \\

\begin{mdframed}[innertopmargin=5pt,linecolor=black,roundcorner=5pt,backgroundcolor=yellow!10]   Summary - The XAI operations have been particularly explored in domains such as healthcare and finance. As AI services become available through pre-trained models bundled with elastic cloud computing resources, the operations of XAI in such a context still require thorough architectural-level research.   
\end{mdframed}

\section{The Context of Early Adoption of XAI Services}
\label{sec:Integration}

The goal of XAI adoption is compatible with the objectives of model quality assurance. We design the XAI operations function as a probe into AI models with or without the model's intrinsic structure to provide explanations, for instance, on how features may affect the learning results. Hence, we propose the early adoption of XAI operations revolves around three major \texttt{core components}:
(1) definition of augmented quality assurance metrics for explanation stability and consistency; 
(2) compatible architecture styles to integrate with cloud AI service development and deployment; 
and (3) XAI operations are configurable and measurable in the same manner as cloud AI services. 
In addition, the adoption of XAI should be cloud-independent and allow cross-validation of multiple XAI methods, AI models, and datasets. 




\subsection{Augmented Metrics for Explanation Quality Assurance}
Several studies \cite{chinu2023explainable,vilone2021notions,Trustworthy} have shown that XAI methods do not always offer consistent explanations, especially in experiments involving Post-hoc XAI methods.
Beyond the model performance, the adoption of XAI operations should measure consistency to ensure model explainability quality.
XAI consistency metrics \cite{xaiprocess}, also shown in Algorithm \ref{eq:stability_metric}, comprise both \texttt{Explanation Stability} and \texttt{Explanation Consistency}. 

\begin{algorithm}
\caption{Calculation of Explanation Metrics \cite{xaiprocess}}
\label{eq:stability_metric}
\begin{algorithmic}[1]
\State \textbf{Input:} Set of explanation summaries \(E = \{\xi^{1}, \xi^{2}, \ldots, \xi^{m}\}\)
\State \textbf{Output:} \(f_d^{K}\) (Stability); \(f_d^{X}\) (Consistency)

\Statex \textit{Notations:}
\Statex \( m \) - Number of summaries in \(E\)
\Statex \( K \) - Combinations, \( \binom{m}{2} \)
\Statex \( \xi^{i} \) - \(i\)-th explanation summary
\Statex \( \xi^{X} \) - Summary for XAI method \(X\)
\Statex \( f_d^{[k]} \) - Prediction changes for \(k\)-th pair

\Procedure{Explanation Stability}{E}
    \State \( K \gets \binom{m}{2} \)
    \State \( f_d^{K} \gets \frac{1}{K} \sum_{k=1}^{K} f_d^{[k]}(\xi^{i},\xi^{j}) \) where \(i \neq j\), \(i,j \leq m\)
\EndProcedure

\Procedure{Explanation Consistency}{E, X}
    \State \( f_d^{X} \gets \frac{1}{m-1} \sum_{k=1}^{m-1} f_d^{[k]}(\xi^{X},\xi^{i}) \) where \(i \leq m\)
\EndProcedure

\end{algorithmic}
\end{algorithm}

\texttt{Explanation Stability} measures the consistency among explanations from many data samples. To compute this metric, we consider all possible pairs of prediction changes from data samples. We then average all these values to a metric, \(f_d^{K}\), representing stability. Theoretically, a smaller value of this metric indicates higher consistency among the explanations generated by the XAI method. We employ the \texttt{Explanation Stability} in the pilot evaluation.
    
\texttt{Explanation Consistency} measures the consistency between different XAI methods. We calculate the prediction changes among XAI methods and average all to \(f_d^{X}\). A smaller value indicates that different XAI methods are producing similar explanations.




\subsection{Compatible Architecture Styles}
The adoption of XAI operations should function in the compatible architecture context of MLOps. As pre-trained ML models on the cloud are available, cloud services become the encapsulation of the models running on elastic computing resources. The communication between XAI methods and AI models thus follows the service orientation.  
We propose \texttt{XAIport}, a service architecture in which the core XAI components are each represented as a microservice with the Open API definition. The APIs for the \texttt{XAIport} service are organized according to Open API 3.0 \cite{swagger2023openapi} standards and documents on SwaggerHub. 

\begin{figure}[ht]
\centerline{\includegraphics[width=\linewidth]{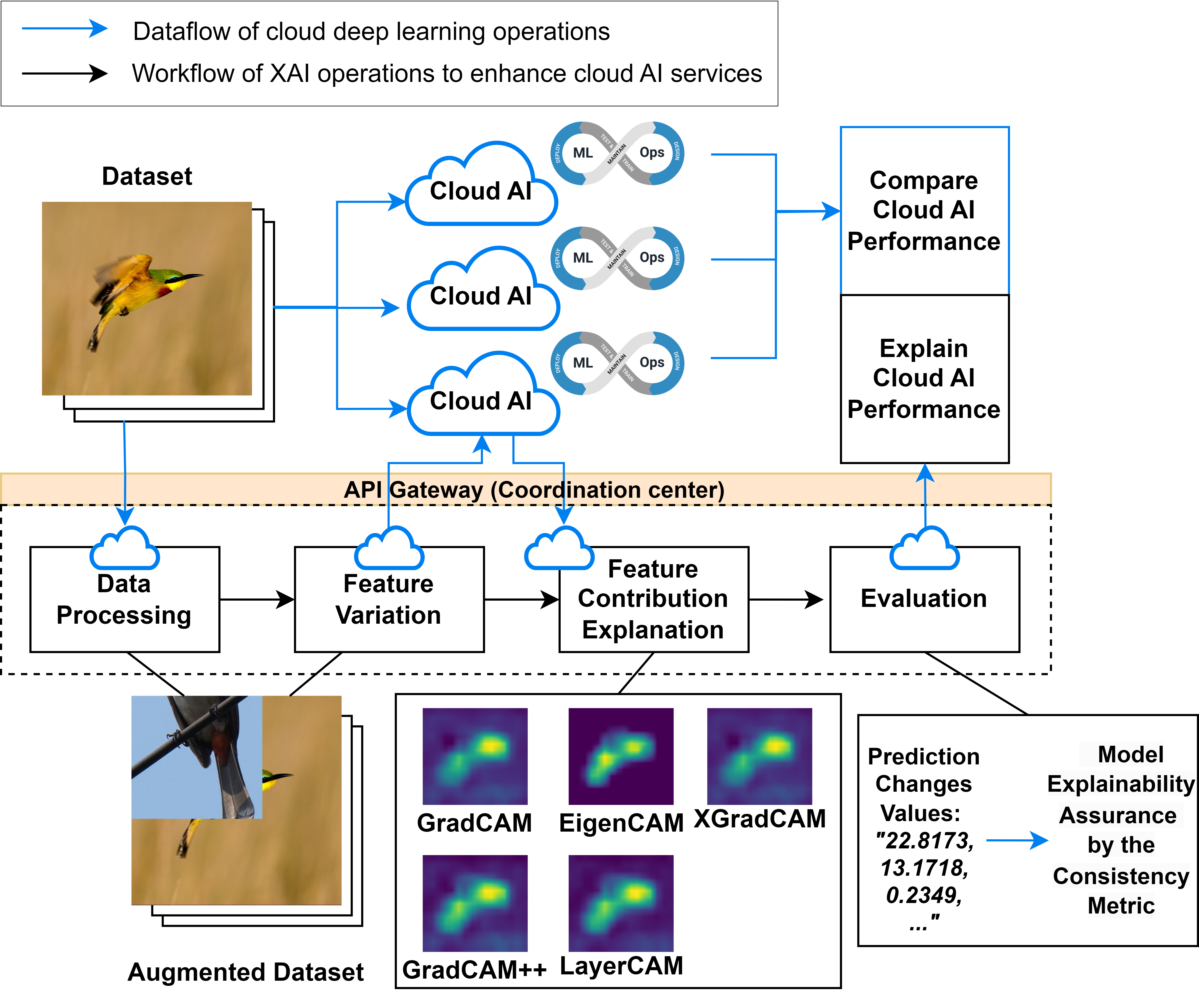}}
\caption{An Illustrating Scenario of Adopting XAI to Multiple Computer Vision Cloud AI Model Development.}
\label{fig:illustration}
\end{figure}

The core architecture includes:  
\texttt{(1) Coordination Center} uses a configuration template to specify pipelines of end-to-end XAI operations from data input, to feature variation, to model inference, to feature contribution explanation, and to evaluation generation; 
\texttt{(2) Data Processing and Storage} is responsible for data preparation and storage intermediate results for explanation generation; 
\texttt{(3) XAI Microservices} encapsulate state-of-the-art XAI methods computing the feature contribution explanation for AI models on a certain dataset; 
\texttt{(4) Evaluation Microservices} computes and visualizes the metrics for XAI explanation. These services produce the answers to questions such as \textit{how is an AI model affected by feature representation?} 
Figure \ref{fig:illustration} provides an illustration of integrating XAI operations along with the development of AI models using cloud AI. We assume the development of AI models adopts the best practices and technology supports from MLOps and DevOps \cite{kreuzberger2023machine,testi2022mlops,hopkins2021machine,DevOps}. The XAI operations are encapsulated as microservices and deployed on the cloud as well. The communication between XAI operations and cloud AI models is only through the endpoints defined by Open APIs. Thus, the enhanced explanation metrics from XAI provide extra measurements for AI model quality assurance. 

\textbf{Integration with Open Community Pretrained Models.} The service-oriented open API architecture \texttt{XAIport} is extensible to the AI model development based on open community libraries such as Hugging face \cite{huggingface}. First, the pre-trained models are accessible for trial and testing with open APIs in the same communication model as the cloud AI services in Figure \ref{fig:illustration}. In addition, when the pre-trained models are downloaded for further retraining and fine-tuning on a domain-specific dataset, the model is deployed in a containerized virtual machine that can be run on a cloud or on a proprietary data center. If we assume such a model is further encapsulated with Open API access, then the \texttt{XAIport} architecture illustrated in Figure \ref{fig:illustration} is applicable without any further changes since \texttt{XAIport} decouples XAI operations and AI models through only the endpoint communication through Open APIs. 

\textbf{Extension to Support A/B Testing of AI Services.} Performing A/B testing on ML models has been adopted by real-world services \cite{larsen2023statistical}. The endpoint of an AI model in the form of Open APIs is the operation unit for automated deployment with multiple production variants for A/B testing. 
As illustrated in Figure ~\ref{fig:illustration}, the XAI services with Open API as endpoints are capable of either covering a certain variant (such as the data augmentation) or communicating with these A/B testing variant endpoints and derive the explanation to link the model's learning performance and feature contributions. 

\subsection{Configurable and Measurable XAI Operations}

Current XAI methods are in the form of algorithms and disparate library code \cite{xaiprocess,adadi2018peeking,lundberg2017unified,arrieta2020explainable}. We propose the XAI operations should be configurable and measurable.
XAI methods, particularly post-hoc techniques, require additional processing power and computational time, thereby increasing the computational overhead \cite{rudin2019stop}. 
 Hardware-wise, the CPU, memory, and GPU requirements may rise to accommodate additional XAI processes. 
 We target to methodically evaluate the operational overhead incurred in the integration and deployment of XAI services across multifarious cloud providers. We focus on measuring the XAI service deployment time and complexity. This exercise entails multiple steps:

\textbf{Selection of CI/CD Tool.} Our methodology commences with the use of cloud pipeline and build tools as the designated continuous integration and continuous deployment (CI/CD) pipeline tool. For instance, Amazon Web Services CodeBuild furnishes essential building facilities for containerized applications, which is imperative for orchestrating a cloud-agnostic XAI milieu, ensuring uniform deployment across diverse cloud platforms.

\textbf{Measuring XAIport Computational Overhead.}
We evaluate the computational overhead of diverse AI models and XAI techniques using \texttt{CodeCarbon} \cite{codecarbon}. This tool, previously applied to several projects \cite{rolnick2022tackling, thirunavukarasu2023large}, is incorporated into \texttt{XAIport} to track time, energy, and carbon footprints during XAI activities and AI predictions. Using \texttt{CodeCarbon} \cite{codecarbon}, we differentiate the energy efficiency and time consumption of various XAI operations, guiding the selection of the optimal method for specific use cases.


\textbf{Measuring XAI Service Deployment Overhead.}
We assess the effort needed to deploy \texttt{XAIport} on multiple cloud providers. 
We perform Amazon Web Services Elastic Container Service (ECS), Azure Virtual Machines, Azure Container Instances, and Google Kubernetes Engine for their efficiency in deploying AI container applications.


\section{A Pilot Evaluation}

We conduct a pilot study using \texttt{XAIport} to explore the answer to a data-driven question for cloud-based AI services as follows. 
\begin{mdframed}[innertopmargin=5pt,linecolor=black,roundcorner=5pt,backgroundcolor=yellow!10] Can early adoption of XAI improve the learning performance of cloud computer vision services? If any, can the improvement be explained? Can the explanation result be evaluated? 
\end{mdframed}


This study uses five visual explanation algorithms, shown in Table \ref{tab:DataAugmentation}, applied to three image classification computer vision services, which are Microsoft Azure Cognitive Services \cite{azure-cognitive-services}, Google Cloud Vertex AI \cite{google_cloud_vertex_ai}, and Amazon Rekognition \cite{aws-ai-services}. 
Cloud platforms offer the following advantages. First, they automate deployment with resource allocation. Second, they offer built-in scalability to manage computational demands efficiently. Third, they deliver monitoring tools for basic model performance. 
However, These platforms overlook the explainability of models.  
In this case, we adopt the \texttt{XAIport} service framework. Upon evaluating the three cloud AI services, we identify potential areas for further optimization in both model performance and explanation stability.
Then, with the integration of Cutmix \cite{cutmix} and Puzzlemix \cite{puzzlemix} data augmentation techniques in the XAI operation, we enhance both the cloud model performance and explanation stability on these platforms.

\subsection{Improving Cloud AI Explanation Metrics with Early XAI Adoption}
We apply the ImageNet dataset \cite{deng2009imagenet} via the \texttt{XAIport} APIs to explore the data-driven question on AI model development. 
The baseline is the cloud AI service trained only using the original dataset.
We adopt five XAI algorithms, which are Grad-CAM \cite{selvaraju2017grad}, Grad-CAM++ \cite{Grad-cam++}, EigenCAM \cite{EigenCAM}, LayerCAM \cite{layercam}, and XGrad-CAM \cite{fu2020axiom}. Adoption of these XAI methods takes image data as inputs from the data processing service and generate saliency maps that highlight the focal areas of layers of the model. Then, these data become the inputs for the three cloud AI and return prediction scores. Finally, we use the evaluation service to derive prediction changes and the stability metrics as algorithm \ref{eq:stability_metric}. 


\begin{table}[ht]
  \caption{Model and XAI Evaluation Results}
  \label{tab:DataAugmentation}
  \begin{adjustbox}{width=\linewidth, center}
  {\Large  
  \begin{tabular}{lccccccc}
    \toprule
    \textbf{Service} & \textbf{F1-score} & \multicolumn{5}{c}{\textbf{XAI Evaluation}} \\ 
    \cmidrule{3-7}
    & & \textbf{GradCAM} & \textbf{GradCAM++} & \textbf{EigenCAM} & \textbf{LayerCAM} & \textbf{XGradCAM} \\
    \midrule
    Azure (B)& 0.839 & 22.227 & 21.211 & 32.498 & 20.595 & 22.229 \\
    Google (B)& 0.565 & 22.329 & 21.233 & 30.713 & 21.328 & 22.327 \\
    Amazon (B)& 0.807 & 18.900 & 17.505 & 30.119 & 17.027 & 18.900 \\
    \midrule
    Azure (C) & 0.864 & 4.544 & 3.773 & 22.072 & 0.339 & 0.339 \\
    Google (C) & 0.876 & 4.316 & 4.437 & 18.427 & 5.147 & 4.316 \\
    Amazon (C) & 0.818 & 13.474 & 11.901 & 28.623 & 12.120 & 13.475 \\
    \midrule
    Azure (P) & 0.905 & 0.078 & 0.107 & 4.732 & 0.002 & 0.002 \\
    Google (P) & 0.869 & 10.440 & 10.246 & 20.754 & 10.781 & 10.440 \\
    Amazon (P) & 0.828 & 14.316 & 13.179 & 26.105 & 2.797 & 3.724 \\
    \bottomrule
  \end{tabular}
  } 
  \end{adjustbox}
  \footnotesize{Note: "B" stands for baseline without augmentation, "C" stands for "Cutmix" and "P" stands for "Puzzlemix". The values in the table are XAI stability metrics \cite{xaiprocess}. (The smaller, the better explanation stability.)}
\end{table}

\textbf{\textit{Results and Discussion.}}
Table \ref{tab:DataAugmentation} shows the detailed measurement results. 
The F1-score shows a subtle enhancement when both CutMix \cite{cutmix} and PuzzleMix \cite{puzzlemix} techniques are employed. A pronounced improvement is observed in the model's explanation evaluation.
The consensus of XAI explanations has alignment with the performance of cloud AI services. 
Such a consistent explanation result provides a trustworthy view of the contribution of the data-driven technique to the model performance improvement. 

These cloud AI services are entirely black-box. There is a lack of access to the model's parameters, the internal network structure, fine-tuned loss functions, and so on. The adoption of XAI through service orientation and open APIs has enabled us to probe the performance and obtain explanations.





\subsection{Computational Analysis of XAI Operations and Deployment Across Cloud Services}

We record and decompose the time spent on (1) Data Processing, (2) Feature Variation, (3) Cloud Inference, (4) XAI and (5) Explanation Stability. We analyze the operation consumption across the three cloud AI services and the five CAM-based XAI methods.
\begin{figure}[ht]
\centering
\includegraphics[width=\linewidth]{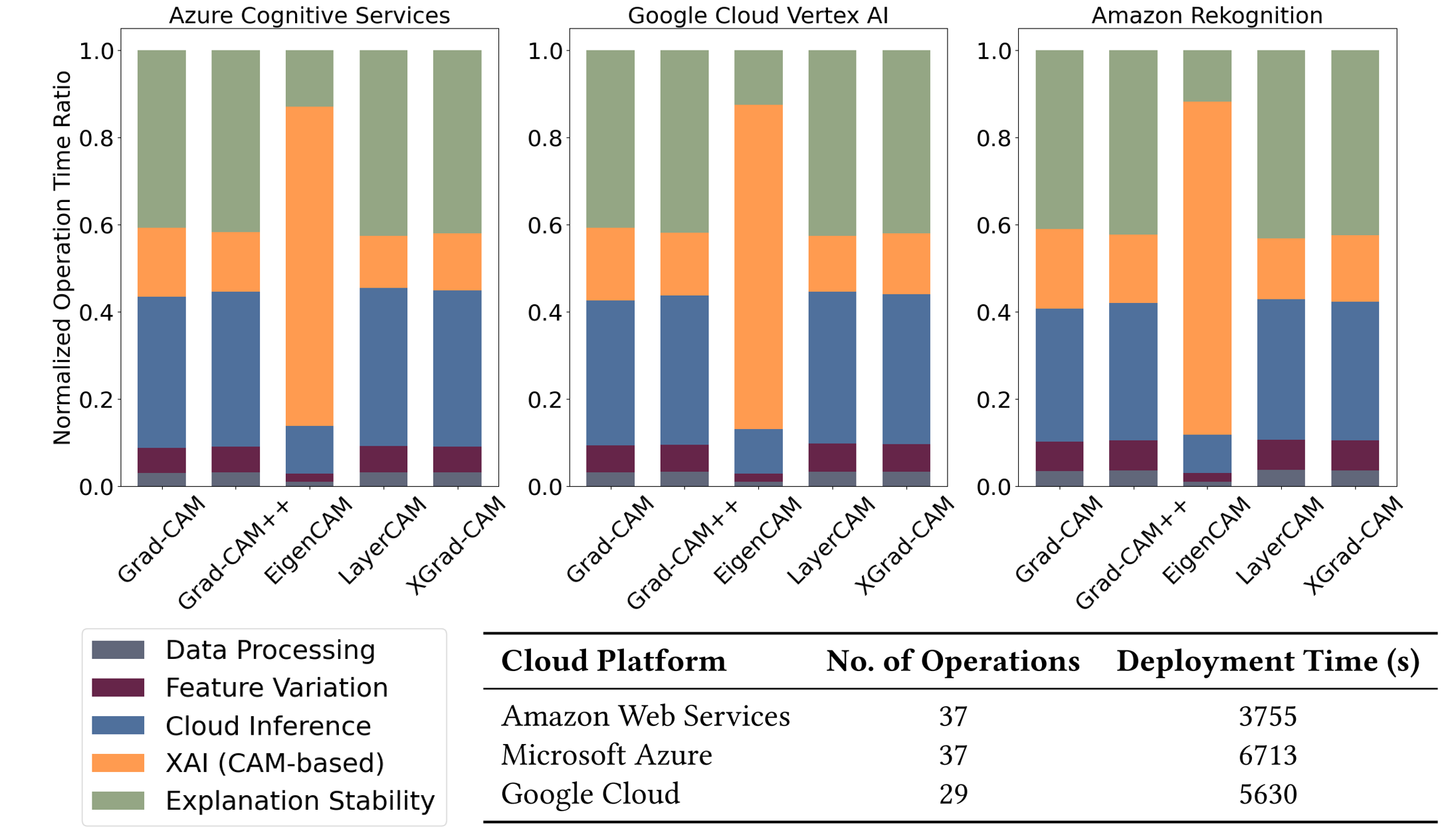}
\parbox{\linewidth}{\footnotesize{Note: The chart displays the decomposition of average XAI execution time per data sample, derived from 1,000 experiments: 
Data Processing (0.12s \(\pm\) 0.03s), Feature Variation (0.23s \(\pm\) 0.06s), Cloud Inference (Azure 1.39s \(\pm\) 0.42s, Google 1.26s \(\pm\) 0.48s, Amazon 1.06s \(\pm\) 0.28s), XAI methods: GradCAM (0.63s \(\pm\) 0.12s), GradCAM++ (0.53s \(\pm\) 0.08s), EigenCAM (9.19s \(\pm\) 3.38s), LayerCAM (0.46s \(\pm\) 0.06s), XgradCAM (0.51s \(\pm\) 0.11s), Explanation Stability (1.60s \(\pm\) 0.56s).}}
\caption{XAI Operations and Framework Deployment Time}
\label{fig:timeratio}
\end{figure}


Figure \ref{fig:timeratio} shows the composition of each unit in XAI operations on average per data sample and the framework deployment duration.
The model inference time is relatively stable across cloud services. 
However, XAI methods take different demands and there is a need for optimization in the evaluation metrics.

\section{Conclusion}
This paper outlines the early adoption of XAI operations in the practices of AI model quality assurance. We define the adoption context in three aspects with mature development methods and technology supports. We illustrate a pilot study on adopting XAI operations to answer a data-driven question with regard to improving three cloud AI services. We demonstrate consistent explanation results with measurements of the computation and deployment overhead. We advocate such a context of practice to broad open AI models' quality assurance with XAI to gain trustworthiness.  In future work, we aim to further develop the software development toolkit (SDK) based on the \texttt{XAIport} framework. The SDK provides the tools for automated deployment of XAI operations along the AI service development using open pre-trained models, dynamic A/B testing of AI services, and validation of new XAI methods.  

\bibliographystyle{ACM-Reference-Format}
\bibliography{reference}

\end{document}